\documentclass{article}

\usepackage{doc}
\usepackage{makeidx}
\usepackage{amsmath}
\usepackage{graphicx}
\usepackage{subcaption}
\usepackage[linesnumbered,lined,boxed,commentsnumbered, ruled, vlined, noend]{algorithm2e}
\usepackage{comment}
\usepackage{authblk}

\usepackage{color}
\definecolor{red}{rgb}{1.00,0.20,0.20}
\definecolor{blue}{rgb}{0.20,0.20,1.00}
\definecolor{green}{rgb}{0.00,1.00,0.00}

\makeatletter
\newlength \fullwidth
\makeatother

\makeatletter
\newlength \figwidth
\makeatother

\makeatletter
\newlength \plotwidth
\makeatother

\makeatletter
\newlength \plotwidthTwo
\makeatother

\makeatletter
\newlength \plotwidthThree
\makeatother

\title{Embedded real-time stereo estimation \\
       via Semi-Global Matching on the GPU}

\author[1]{D. Hernandez-Juarez}
\author[1]{A. Chac{\'o}n}
\author[1]{A. Espinosa}
\author[2]{D. V{\'a}zquez}
\author[1]{\\J. C. Moure}
\author[2]{A. M. L{\'o}pez}
\affil[1]{Computer Architecture \& Operating Systems Department (CAOS) at Universitat Autonoma de Barcelona}
\affil[2]{Computer Vision Center}

\begin{document}

\maketitle


\begin{abstract}
Dense, robust and real-time computation of depth information from stereo-camera systems is a computationally demanding requirement for robotics, advanced driver assistance systems (ADAS) and autonomous vehicles. Semi-Global Matching (SGM) is a widely used algorithm that propagates consistency constraints along several paths across the image. This work presents a real-time system producing reliable disparity estimation results on the new embedded energy-efficient GPU devices. Our design runs on a Tegra X1 at 42 frames per second (fps) for an image size of 640$\times$480, 128 disparity levels, and using 4 path directions for the SGM method.
\end{abstract}

\section{Introduction}
\label{sect:introduction}

Dense, robust and real-time computation of depth information from stereo-camera systems is a requirement in many industrial applications such as advanced driver assistance systems (ADAS), robotics navigation and autonomous vehicles. An efficient stereo algorithm has been a research topic for decades \cite{arabnia1995distributed}. It has multiple applications, for example, \cite{Gonzalez:2015} uses stereo information to filter candidate windows for pedestrian detection and provides better accuracy and performance.

\begin{figure}[t]
\centering
\includegraphics[width=0.92\textwidth]{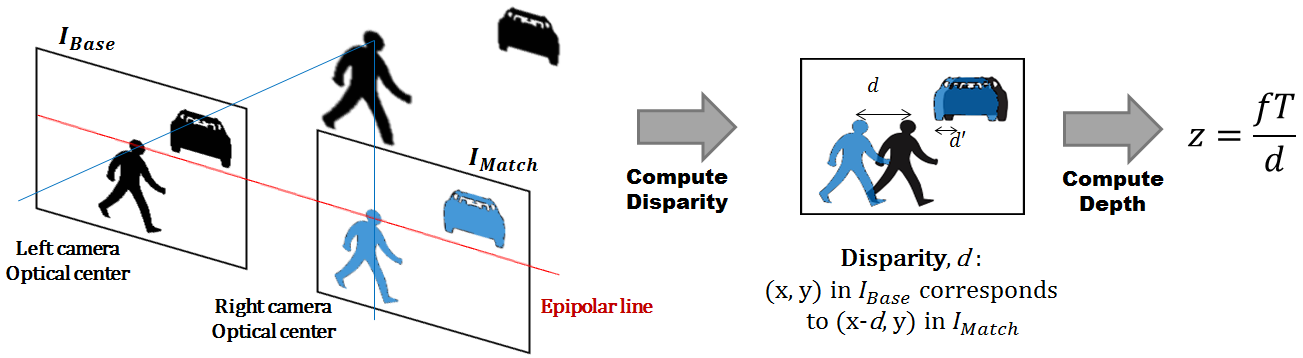}
\caption{Each pixel of $I_{Base}$ corresponds to one pixel of $I_{Match}$, and the epipolar geometry of the two cameras limits the search to a one dimensional line. The distance $z$ between the 3D point and the baseline of the camera is computed from the disparity $d$ using triangulation, where $f$ is the focal length and $T$ is the baseline of the camera pair.
}
\label{fig:cameras}
\end{figure}

Fig. \ref{fig:cameras} illustrates how to infer the depth of a given real-world point from its projection points on the left and right images. Assuming a simple translation between the cameras (otherwise, images must be rectified using multiple extrinsic and intrinsic camera parameters), the corresponding points must be in the same row of both images, along the epipolar lines. A similarity measure correlates matching pixels and the $disparity$ ($d$) is the similarity distance between both points.

Disparity estimation is a difficult task because of the high level of ambiguity that often appears in real situations. For those, a large variety of proposals have been extensively presented \cite{scharstein2002taxonomy}. Most of the high-accuracy stereo vision pipelines \cite{vzbontar2014computing} include the semi-global matching (SGM) consistency-constraining algorithm \cite{hirschmuller2008stereo}. The combination of SGM with different kinds of local similarity metrics is insensitive to various types of noise and interferences (like lighting), efficiently deals with large untextured areas and is capable of retaining edges.

The high computational load and memory bandwidth requirements of SGM pose hard challenges for fast and low energy-consumption implementations. Dedicated hardware solutions (e.g. FPGA or ASIC) \cite{banz2010real}\cite{paya2010vliw} achieve these goals, but they are very inflexible regarding changes in the algorithms. Implementations on desktop GPUs can assure real-time constraints \cite{banz2011real}, but their high power consumption and the need to attach a desktop computer makes them less suitable for embedded systems.

Recently, with the appearance of embedded GPU-accelerated systems like the NVIDIA Jetson TX1 and the DrivePX platforms (incorporating, respectively, one and two Tegra X1 ARM processors), low-cost and low-consumption real-time stereo computation is becoming attainable. The objective of this work is to implement and evaluate a complete disparity estimation pipeline on this embedded GPU-accelerated device. 

We present simple, but well-designed, baseline massively parallel schemes and data layouts of each of the algorithms required for disparity estimation, and then optimize the baseline code with specific strategies, like vectorization or $CTA$-$to$-$Warp$ conversion, to boost performance around 3 times. The optimized implementation runs on a single Tegra X1 at 42 frames per second (fps) for an image size of 640$\times$480 pixels, 128 disparity levels, and using 4 path directions for the SGM method, providing high-quality real-time operation. While a high-end desktop GPU improves around 10 times the performance of the embedded GPU, the performance per watt ratio is 2.2 times worse. The source code is available\footnote{https://github.com/dhernandez0/sgm}.

The rest of the paper is organized as follows. Section \ref{sect:disparity_pipeline} presents the algorithms composing the disparity estimation pipeline, overviews the GPU architecture and programming model and discusses related work. In section \ref{sect:parallelization} we describe each algorithm and then propose and discuss a parallel scheme and data layout. Finally, section \ref{sect:results} provides the obtained results and section \ref{sect:conclusions} summarizes the work.

\section{Disparity Estimation Pipeline}
\label{sect:disparity_pipeline}

\begin{figure}[t]
\centering
\includegraphics[width=1.0\textwidth]{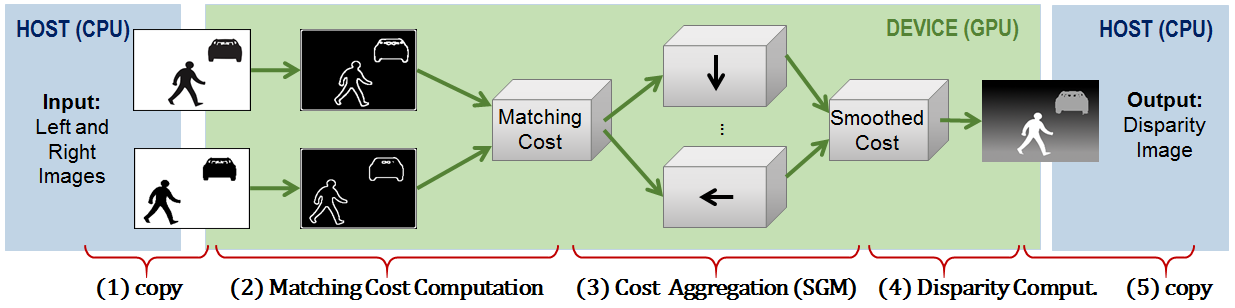}
\caption{Stages of the GPU-accelerated Disparity Estimation Pipeline}
\label{fig:pipeline}
\end{figure}

Fig. \ref{fig:pipeline} shows the stages of the disparity computation pipeline: (1) the captured images are copied from the Host memory space to the GPU Device; (2) features are extracted from each image and used for similarity comparison to generate a local matching cost for each pixel and potential disparity; (3) a smoothing cost is aggregated to reduce errors (SGM); (4) disparity is computed and a 3$\times$3 median filter is applied to remove outliers; and (5) the resulting disparity image is copied to the Host memory.

\subsection{Local Matching Cost and Semi-Global Matching (SGM)}

Different similarity metrics or cost functions have been proposed in the literature. The less computationally-demanding, and modest quality providers, are Sum of Absolute Differences, ZSAD and Rank Transform. According to \cite{hirschmuller2009evaluation}, Hierarchical Mutual Information and the Census Transform (CT) features \cite{zabih1994non} provide similar higher quality, being CT substantially less time-consuming. Recently, costs based on neural networks have outperformed CT \cite{vzbontar2014computing}, but at the expense of a higher computational load.

A CT feature encodes the comparisons between the values of the pixels in a window around a central pixel. After empirically evaluating different variants we selected a Center-Symmetric Census Transform (CSCT) configuration with a 9$\times$7 window, which provides a more compact representation with similar accuracy \cite{spangenberg2013weighted}. The similarity of two pixels is defined as the Hamming distance of their CSCT bit-vector features. Two properties provide robustness for outdoor environments with uncontrolled lighting and in front of calibration errors: the invariance to local intensity changes (neighboring pixels are compared to each other) and the tolerance to outliers (an incorrect value modifies a single bit).


In order to deal with non-unique or wrong correspondences due to low texture and ambiguity, consistency constraints can be included in the form of a global two-dimensional energy minimization problem. Semi-global matching (SGM) approximates the global solution by solving a one-dimensional minimization problem along several (typically 4 or 8) independent paths across the image. For each path direction, image point and disparity, SGM aggregates a cost that considers the cost of neighboring points and disparities. The number of paths affects both the quality and the performance of the results.

\subsection{Overview of GPU architecture and performance}
\label{sect:gpu}


GPUs are massively parallel devices containing tens of throughput-oriented processing units called \emph{streaming multiprocessors} (SMs). Memory and compute operations are executed as vector instructions and are highly pipelined in order to save energy and transistor budged. SMs can execute several vector instructions per cycle, selected from multiple independent execution flows: the higher the available parallelism the better the pipeline utilization.

The CUDA programming model allows defining a massive number of potentially concurrent execution instances (called $threads$) of the same program code. A unique two-level identifier \textless $ThrId$, $CTAid$\textgreater $ $ is used to specialize each thread for a particular data and/or function. A CTA ($Cooperative$ $Thread$ $Array$) comprises all the threads with the same $CTAid$, which run simultaneously and until completion in the same SM, and can share a fast but limited memory space. $Warps$ are groups of threads with consecutive $ThrId$s in the same CTA that are mapped by the compiler to vector instructions and, therefore, advance their execution in a lockstep synchronous way. The warps belonging to the same CTA can synchronize using a explicit barrier instruction. Each thread has its own private local memory space (commonly assigned to registers by the compiler), while a large space of global memory is public to all execution instances (mapped into a large-capacity but long-latency device memory, which is accelerated using a two-level hierarchy of cache memories).

The parallelization scheme of an algorithm and the data layout determine the available parallelism at the instruction and thread level (required for achieving full resource usage) and the memory access pattern. GPUs achieve efficient memory performance when the set of addresses generated by a warp refer to consecutive positions that can be \emph{coalesced} into a single, wider memory transaction. Since the bandwidth of the device memory can be a performance bottleneck, an efficient CUDA code should promote data reuse on shared memory and registers.

\subsection{Related work}
\label{sect:related_work}

A reference implementation of SGM on CPU \cite{spangenberg2014large} reached 5.43 frames per second (fps) with 640$\times$480 image resolution and 128 disparity levels. They applied SGM with 8 path directions ad an additional left-right consistency check and sub-pixel interpolation. A modified version with reduced disparity computation (rSGM) was able to reach 12 fps.


Early GPU implementations \cite{ernst2008mutual} and \cite{rosenberg2006real} present OpenGL/Cg SGM implementations with very similar performance results peaking at 8 fps on 320$\times$240 resolution images.Versions designed for early CUDA systems and proposed specific modifications of the SGM algorithm. Haller and Nedevschi \cite{haller2010gpu} modified the original cost aggregation formula removing the P1 penalty and using 4 path directions for cost aggregation. In this way, they reduced computation and memory usage, but also reduced accuracy. Their implementation reached 53 fps on a Nvidia GTX 280 with images of 512$\times$383. 

The most recent implementation \cite{banz2011real} stated very fast results: 27 fps on 1024$\times$768 images using a NVIDIA Tesla C2050, with 128 disparity levels. By using Rank Transform \cite{zabih1994non} as matching cost function, their proposal provides lower accuracy \cite{hirschmuller2009evaluation}. We will notice some differences in the parallel scheme on the following discussion.

As far as we know this is the first evaluation of disparity estimation in a Nvidia GPU-accelerated embedded system, as well as in the last Maxwell architecture. We propose better parallelization schemes to take advantage of the hardware features available in current systems. 



\section{Algorithm Description and Massive Parallelization}
\label{sect:parallelization}

This section describes the algorithms used for disparity computation and discusses the alternative parallelization schemes and data layouts. We present the baseline pseudocode for the proposed massively parallel algorithms and explain additional optimizations.

\subsection {Matching Cost Computation}
\label{sect:census_transform}

A 9$\times$7-window, Center-Symmetric Census Transform (CSCT) concatenates the comparisons of 31 pairs of pixels into a bit-vector feature. Equation \ref{eq:census_transform} defines the CSCT, where $\otimes$ is bit-wise concatenation, $I(x,y)$ is the value of pixel ($x$,$y$) in the input image, and $s$($u$,$v$) is $1$ if $u$ $\geq$ $v$, or $0$ otherwise. The matching cost $MC(x,y,d)$ between a pixel ($x,y$) in the base image and each potentially corresponding pixel in the match image at disparity $d$ is defined by equation \ref{eq:matching_cost}, where $\oplus$ is bit-wise exclusive-or and $bitcount$ counts the number of bits set to 1.

\begin{equation}
CSCT_{9,7}(I,x,y) = \bigotimes \begin{cases}
			\bigotimes\limits_{i=1}^{4} \bigotimes\limits_{j=-3}^{3} s( I(x+i,y+j), I(x-i, y-j))\\
			\bigotimes\limits_{j=1}^{3} s( I(x,y+j), I(x, y-j))
		\end{cases}
\label{eq:census_transform}
\end{equation}

\begin{equation}
MC(x, y, d) = bitcount ( CSCT_{9,7}(I_{base}, x, y) \oplus CSCT_{9,7}(I_{match}, x-d, y) )
\label{eq:matching_cost}
\end{equation}

\begin{figure}[b]
\centering
\includegraphics[width=1.0\textwidth]{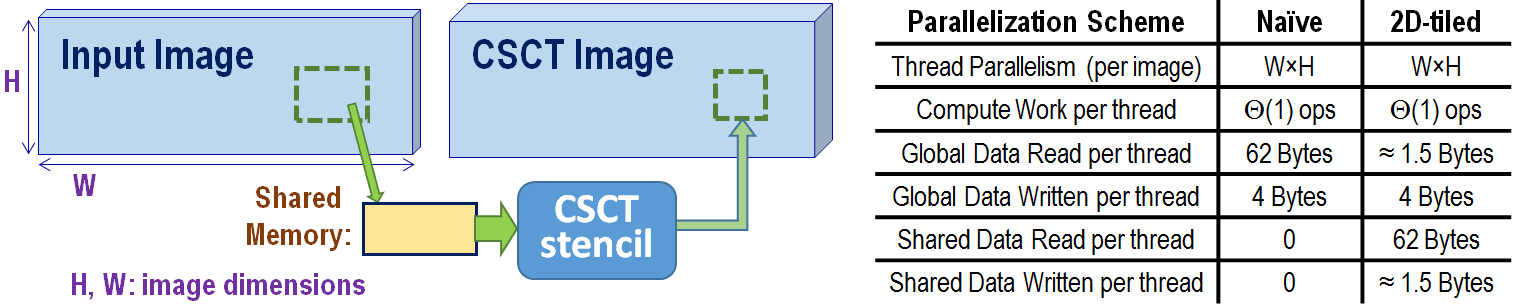}
\caption{CSCT: 2D-tiled CTA-parallel scheme and computational analysis}
\label{fig:csct_comp}
\end{figure}

The data access patterns inherent in both equations exhibit different data reuse schemes, which prevent both algorithms to be fused. The 2D-tiled parallel scheme shown in Fig. \ref{fig:csct_comp} matches the 2D-stencil computation pattern of CSCT, and maximizes data reuse: the attached table shows how a tiled scheme using shared memory reduces the total global data accesses by $(62+4)/(1.5+4) = 12$ times with respect to a straightforward, na\"{i}ve, embarrassingly parallel design, where each thread reads its input values directly from global memory.

\begin{figure}[!b]
\centering
\includegraphics[width=1.0\textwidth]{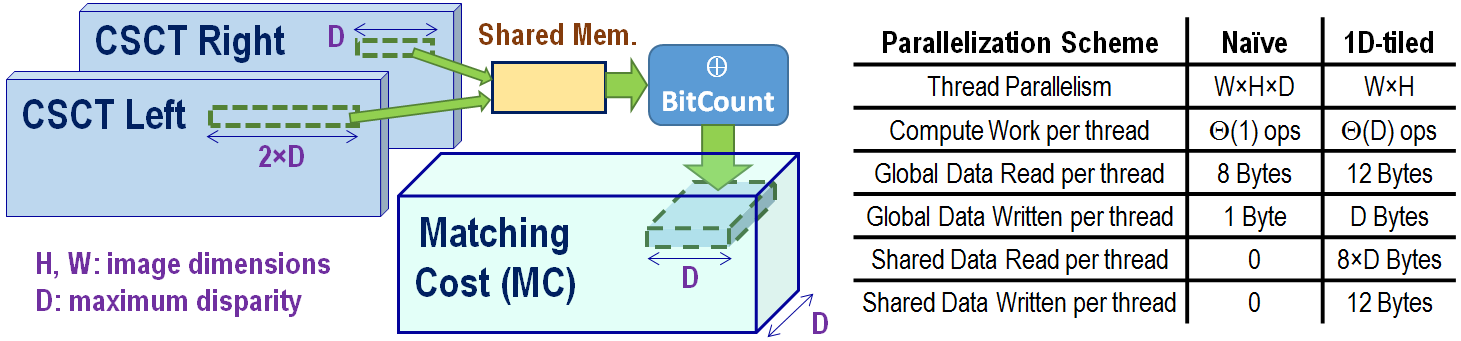}
\caption{Matching cost: 1D-tiled CTA-parallel scheme and computational analysis}
\label{fig:match_cost}
\end{figure}

The 1D-tiled parallel scheme for computing matching cost (MC) exploits data reuse on the x-dimension (see Fig. \ref{fig:match_cost}). As proposed in \cite{banz2011real}, we can represent matching cost using a single byte without losing accuracy, which reduces 4 times the memory bandwidth requirements in comparison to using 32-bit integers. The attached table shows that the read-cooperative scheme, compared to the na\"{i}ve design, sacrifices parallelism (divides the number of threads by $D$, the maximum disparity considered) by higher data reuse (around 8 times less global memory accesses). The low arithmetic intensity of the algorithm (2 main compute operations every 9-Byte memory accesses) advises for this kind of optimization.

Algorithms \ref{alg:CSCT} and \ref{alg:MC} show the pseudocode of both parallel algorithms, not including special code for corner cases handling image and CTA boundaries. In both cases, threads in the same CTA cooperate to read an input data tile into shared memory, then synchronize, and finally perform the assigned task reading the input data from shared memory. The first algorithm assumes a CTA size of $WarpSize$$\times$$WarpSize$ threads and the second algorithm a CTA of $D$ threads. They are both scalable designs that use a small constant amount of shared memory per thread (1.5 and 12 Bytes, respectively).

There are two memory-efficient layout alternatives for algorithm \ref{alg:MC}. Each CTA generates a $D$$\times$$D$ slice in the y-plane of the MC matrix, and threads can generate together the cost for (1) all the disparity levels for the same pixel or (2) all the pixels in the block for the same disparity level. We chose the first option, and adapt the data layout so that the indexes of disparity levels vary faster on the MC cube and global write instructions are coalesced. The second solution, used in \cite{banz2011real}, provides similar performance on this algorithm but compromises the available parallelism and the performance of the following SGM algorithm. 

\begin{algorithm}[!t]
 \SetAlgoLined
 \SetKwData{Left}{left}\SetKwData{This}{this}\SetKwData{Up}{up}
 \SetKwFunction{Union}{Union}\SetKwFunction{FindCompress}{FindCompress}
 \SetKwInOut{Input}{input}\SetKwInOut{Output}{output}
 
\Input{I[$H$][$W$], $H$, $W$}
\Output{CSCT[$H$][$W$]}
 
\textbf{parallel} \For {$y$=$0$ \textbf{to} $H$ \textbf{step} $WarpSize$} {
\textbf{parallel} \For {$x$=$0$ \textbf{to} $W$ \textbf{step} $WarpSize$} {
\textbf{CTA parallel} \For {$yCTA,xCTA$=($0$,$0$) \textbf{to} ($WarpSize$,$WarpSize$)} {
     copy ($WarpSize+8$)$\times$($WarpSize+6$) tile of I[][] into SharedI[][]\;
     CTA Barrier Synchronization\;
     CSCT[$y$+$yCTA$][$x$+$xCTA$] = CSCT$_{9,7}$($SharedI$, $xCTA$, $yCTA$);
   }
  }
 }
\caption{CSCT: 2D-tiled, read-cooperative parallel scheme}
\label{alg:CSCT}
\end{algorithm}

\vspace{-0.2cm}

\begin{algorithm}[t]
 \SetAlgoLined
 \SetKwData{Left}{left}\SetKwData{This}{this}\SetKwData{Up}{up}
 \SetKwFunction{Union}{Union}\SetKwFunction{FindCompress}{FindCompress}
 \SetKwInOut{Input}{input}\SetKwInOut{Output}{output}
 
 \Input{CSCTbase[$H$][$W$], CSCTmatch[$H$][$W$], $H$, $W$, $D$}
 \Output{MC[$H$][$W$][$D$]}
 \BlankLine
\textbf{parallel} \For {$y$=$0$ \textbf{to} $H$ } {
\textbf{parallel} \For {$x$=$0$ \textbf{to} $W$ \textbf{step} $D$} {
   \textbf{CTA parallel} \For {$ThrId$=0 \textbf{to} $D$}{
     SharedM[$ThrId$] =   CSCTmatch[$y$][$x$+$ThrId$-$D$]\;
     SharedM[$D$+$ThrId$] =   CSCTmatch[$y$][$x$+$ThrId$]\;
     SharedB[$ThrId$] =   CSCTbase[$y$][$x$+$ThrId$]\;
     CTA Barrier Synchronization\;
     \For {$i$=$0$ \textbf{to} $D$} {
       MC[$y$][$x$+$i$][$ThrId$] = BitCount ( SharedB[$i$] $\oplus$ SharedM[$ThrId$+$1$+$i$] )\;
}
}
}
}   
\caption{Matching Cost computation: 1D-tiled, read-cooperative parallel scheme; Data layout: MC[$y$][$x$][$d$] ($d$ indexes vary faster) }
\label{alg:MC}
\end{algorithm}

\subsection{Smoothing Cost Aggregation (SGM) and Disparity Computation}
\label{sect:cost_aggregation}

The SGM method solves a one-dimensional minimization problem along different paths $r$=($r_{x},r_{y}$) using the recurrence defined by equation \ref{eq:cost_aggregation} and a dynamic programming algorithmic pattern. Matrix L$_r$ contains the smoothing aggregated costs for path $r$. The first term of equation \ref{eq:cost_aggregation} is the original matching cost, and the second term adds the minimum cost of the disparities corresponding to the previous pixel ($x-r_x$,$y-r_y$), including penalties for small disparity changes ($P_1$) and for larger disparity discontinuities and ($P_2$). $P_1$ is intended to detect slanted and curved surfaces, while $P_2$ smooths the results and makes abrupt changes difficult. The last term  ensures that aggregated costs are bounded. For a detailed discussion refer to \cite{hirschmuller2008stereo}.
The different L$_r$ matrices must be added together to generate a final cost and then select the disparity corresponding to the minimum ($winner$-$takes$-$all$ strategy), as shown by equation \ref{eq:smoothed_costs}.

\begin{equation}
L_r(x,y,d)= MC(x,y,d) + min
\begin{cases}
			L_r(x-r_x, y-r_y, d)\\
			L_r(x-r_x, y-r_y, d-1) + P_1\\
			L_r(x-r_x, y-r_y, d+1) + P_1\\
			min_i L_r(x-r_x, y-r_y, i) + P_2
\end{cases}- min_k L_r(x-r_x, y-r_y,k)
\label{eq:cost_aggregation}
\end{equation}

\begin{equation}
D(x,y) = min_d \sum\limits_{r} L_r(x,y,d)
\label{eq:smoothed_costs}
\end{equation}

Equation \ref{eq:cost_aggregation} determines a recurrent dependence that prevents the parallel processing of pixels in the same path direction. Parallelism can be exploited, though, in the direction perpendicular to the path, in the disparity dimension, and for each of the computed path directions. Our proposal exploits all the available parallelism by creating a CTA for each slice in the aggregated cost matrix along each particular path direction.

\begin{figure}[!b]
\centering
\includegraphics[width=1.0\textwidth]{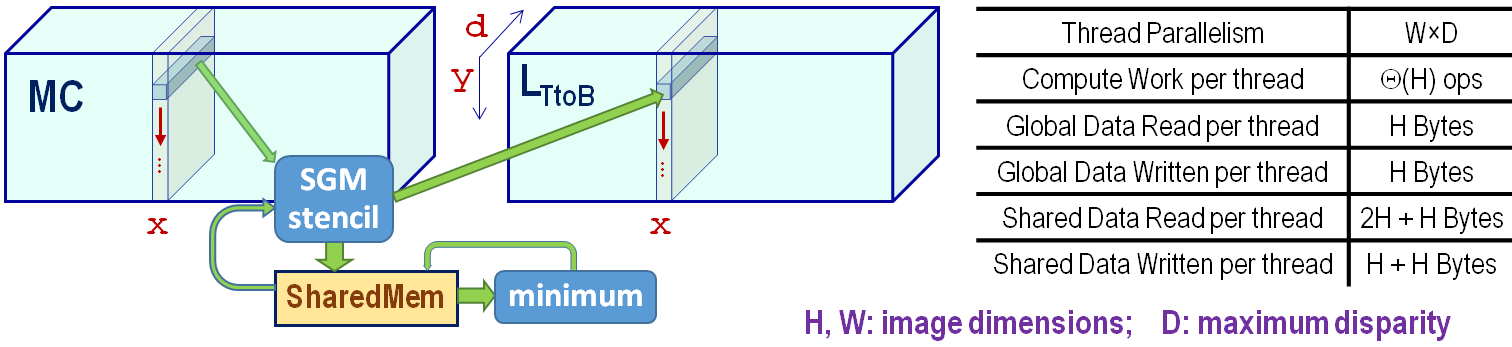}
\caption{Aggregated cost, Top-to-Bottom: CTA-parallel scheme with recurrence in the y-dimension and computational analysis}
\label{fig:aggr_cost}
\end{figure}

Fig. \ref{fig:aggr_cost} illustrates the case of the top-to-bottom path direction and algorithm \ref{alg:Aggr} shows the pseudocode. Each of the $W$ slices is computed by a different CTA of $D$ threads, with each thread executing a recurrent loop (line 4) to generate $H$ cost values along the path. Computing the cost for the current pixel and disparity level requires the cost of the previous pixel on neighboring disparity levels: one value can be reused in a private thread register but the neighboring costs must be communicated among threads (lines 7,8 and 12). Finally, all threads in the CTA must collaborate to compute the minimum cost for all disparity levels (line 11).

The case for horizontal paths is very similar, with $H$ slices computed in parallel. Diagonal path directions are a little more complex: $W$ independent CTAs process the diagonal slices moving in a vertical direction (assuming $W$ $\geq$ $H$). When a CTA reaches a boundary, it continues on the other boundary. For example, a top-to-bottom and right-to-left diagonal slice starting at (x,y) = (100,0) will successively process pixels (99,1), (98,2) ... (0, 100), and then will reset the costs corresponding to the previous pixel and continue with pixels ($W$-1,101), ($W$-2,102) ... 

\begin{algorithm}[!t]
 \SetAlgoLined
 \SetKwData{Left}{left}\SetKwData{This}{this}\SetKwData{Up}{up}
 \SetKwFunction{Union}{Union}\SetKwFunction{FindCompress}{FindCompress}
 \SetKwInOut{Input}{input}\SetKwInOut{Output}{output}
 
 \Input{MC[$H$][$W$][$D$], $H$, $W$, $D$}
 \Output{L[$H$][$W$][$D$]}
 \BlankLine
 \textbf{parallel} \For {x=$0$ \textbf{to} $W$} {
   \textbf{CTA parallel} \For {ThrId=0 \textbf{to} $D$}{
     Initialize $aggr$, $min$ and SharedAggr[] with MAX\_VALUE\;
     \For {y=$0$ \textbf{to} $H$} {
       $cost$ = MC[$y$][$x$][$ThrId$]\;
       CTA Barrier Synchronization\;
       $left$ = SharedAggr[$ThrId$]\;
       $right$= SharedAggr[$ThrId$+2]\;
       $aggr$ = $cost$ + minimum( $aggr$, $left$+P1, $right$+P1, $min$+P2 ) - $min$\;
       L[$y$][$x$][$ThrId$] = $aggr$\;
       $min$ = CTA\_Minimum\_Reduce ( $aggr$ );  *** includes Barrier Synchronization
       SharedAggr[$ThrId$+1] = $aggr$\;
     }
   }
 }
\caption{Aggregated Cost computation: top-to-bottom path direction}
\label{alg:Aggr}
\end{algorithm}


The cost aggregation and disparity computation defined by equation \ref{eq:smoothed_costs} have been fused in Algorithm \ref{alg:DispComputation} in order to reduce the amount of memory accesses (avoids writing and then reading the final cost matrix). A CTA-based parallel scheme is proposed so that each CTA produces the disparity of a single pixel (line 7): first, each CTA thread adds the costs corresponding to a given disparity level for all path directions (line 4), and then CTA threads cooperate to find the disparity level with minimum cost (line 5). 

\begin{algorithm}[!t]
 \SetAlgoLined
 \SetKwData{Left}{left}\SetKwData{This}{this}\SetKwData{Up}{up}
 \SetKwFunction{Union}{Union}\SetKwFunction{FindCompress}{FindCompress}
 \SetKwInOut{Input}{input}\SetKwInOut{Output}{output}
 
\Input{L$_0$[$W$][$H$][$D$], L$_1$[$W$][$H$][$D$], L$_2$[$W$][$H$][$D$] ... $W$, $H$, $D$}
\Output{Disp[$W$][$H$]}

\BlankLine
\textbf{parallel} \For {$x$=$0$ \textbf{to} $W$} {
\textbf{parallel} \For {$y$=$0$ \textbf{to} $H$} {
   \textbf{CTA parallel} \For {ThrId=0 \textbf{to} $D$}{
     $cost$ = L$_0$[$x$][$y$][$ThrId$]+L$_1$[$x$][$y$][$ThrId$]+L$_2$[$x$][$y$][$ThrId$]+...\;
     $MinIndex$ = CTA\_Minimum\_Reduce($cost$, $ThrId$)\;
     \uIf {ThrId == 0} {
       Disp[$x$][$y$] = $MinIndex$\;
      }
    }
   }
 }   

\caption{Summation of all path costs and Disparity Computation}
\label{alg:DispComputation}
\end{algorithm}

\vspace{-0.4cm}
\subsection{Additional Optimizations}
\vspace{-0.1cm}

We have applied three types of optimizations to the baseline algorithms that provided a combined performance improvement of almost 3$\times$. We have vectorized the inner loop of algorithm \ref{alg:Aggr} (lines 4-12) to process a vector of 4 cost values (4 bytes) per instruction (requiring a special byte-wise SIMD instructions for computing the minimum operation). We have also modified the parallel scheme so that a single warp performs the task previously assigned to a CTA, which we call $CTA$-$to$-$warp$ conversion. It (1) avoids expensive synchronization operations, (2) allows using fast register-to-register communication (using special shuffle instructions) instead of shared-memory communications, and (3) reduces instruction count and increases instruction-level parallelism. A drawback of both strategies is a reduction of thread-level parallelism, as shown in \cite{chacon2014thread}. This is not a severe problem in the embedded Tegra X1 device, with a maximum occupancy of $\approx$ 4 thousand threads.

Finally, to reduce the amount of data accessed from memory, the computation of the aggregated cost for the last path direction (Alg. \ref{alg:Aggr} Bottom-to-Top) is fused with the final cost summation and disparity computation (Alg. \ref{alg:DispComputation}), providing a 1.35x performance speedup on the Tegra X1. Also, fusing the computation of the initial matching cost (Alg. \ref{alg:MC}) with the aggregate cost computation for the horizontal path directions (Alg. \ref{alg:Aggr}) improves performance by 1.13x. 




\vspace{-0.1cm}
\section{Results}
\vspace{-0.1cm}
\label{sect:results}


We have measured execution time and disparity estimation accuracy for multiple images, 128 disparity levels, and 2, 4 and 8 path directions. Apart from executing on a NVIDIA Tegra X1, which integrates 8 ARM cores and 2 Maxwell SMs with a TDP of 10W, and for comparison purposes, we have also executed on a high-end NVIDIA Titan X, with 24 Maxwell SMs and a TDP of 250W. We ignore the time for CPU-GPU data transfers (less than 0.5\% of the total elapsed time) since it can be overlapped with computation. Since performance scales proportional to the number of image pixels, we will restrict our explanation to $640$$\times$$480$ images. 

\begin{figure}[!t]
\includegraphics[width=\linewidth]{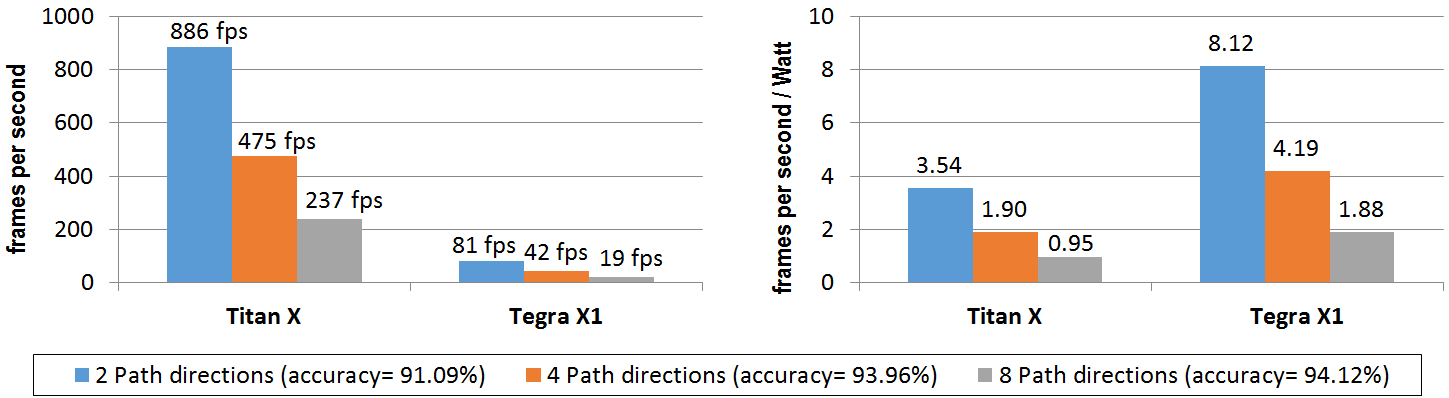}
\caption{Performance (fps), performance per Watt and accuracy results for 640$\times$480px images, 128 disparity levels, and 2, 4 and 8 SGM path directions}
\label{fig:tegra_performance}
\end{figure}

The legend in Fig. \ref{fig:tegra_performance} indicates the disparity estimation accuracy, measured using the KITTI benchmark-suite \cite{Geiger2012CVPR}, when using different SGM configurations, and not considering occluded pixels and treating more than 3 pixel differences as errors. Using 4 path directions (excluding diagonals) reduces accuracy very slightly, while using only the left-to-right and top-to-bottom directions reduces accuracy more noticeably. 

The left and right charts in Fig. \ref{fig:tegra_performance} show, respectively, the performance throughput (frames per second, or fps) and the performance per watt (fps/W) on both GPU systems and also for different SGM configurations. The high-end GPU always provides more than 10 times the performance of the embedded GPU (as expected by the difference in number of SMs), but the latter offers around 2 times more performance per Watt. It is remarkable that real-time rates (42 fps) with high accuracy are achieved by the Tegra X1 when using 4 path directions.

\begin{figure}[h]
\begin{subfigure}{0.49\textwidth}
\includegraphics[width=\linewidth]{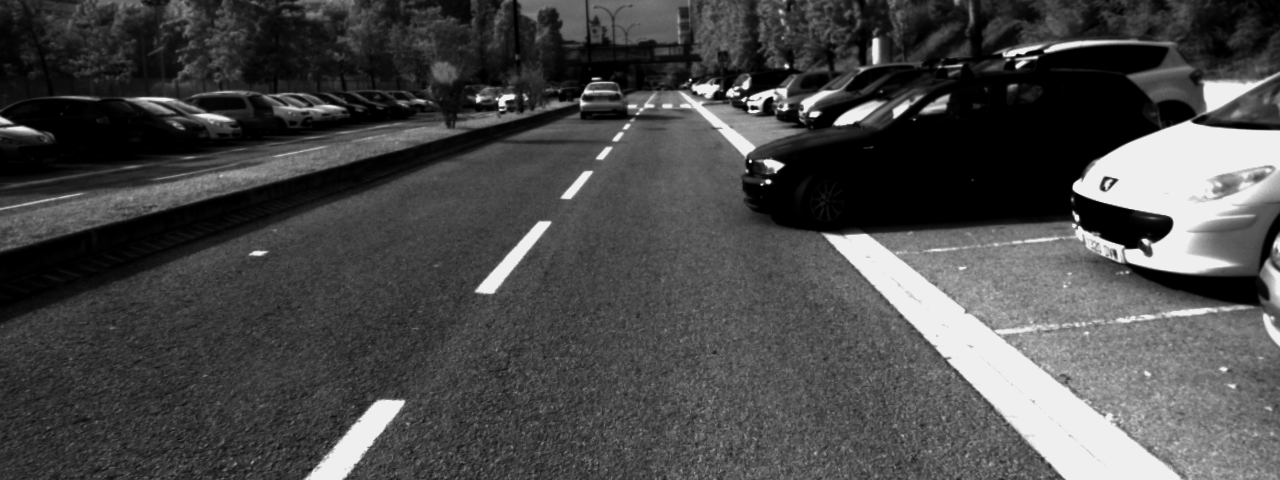}
\caption{Image obtained from the left camera of the car} 
\label{fig:disp_left}
\end{subfigure}
\begin{subfigure}{0.49\textwidth}
\includegraphics[width=\linewidth]{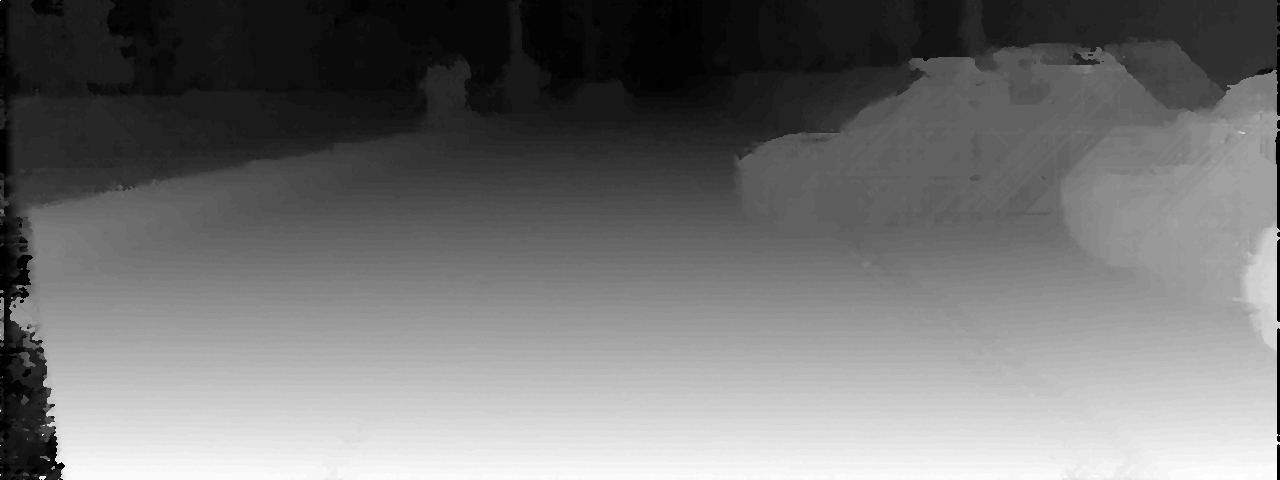}
\caption{Disparity computed with SGM described here}
\label{fig:disp}
\end{subfigure}
\caption{Example of disparity computation}
\end{figure}

Finally, an example of the disparity computed by our proposed algorithm can be seen in Fig. \ref{fig:disp}.

\section{Conclusions}
\label{sect:conclusions}

The results obtained show that our implementation of depth computation for stereo-camera systems is able to reach real-time performance on a Tegra X1. This fact indicates that low-consumption embedded GPU systems, like the Tegra X1, are well capable of attaining real-time processing demands. Hence, their low-power envelope and remarkable performance make them good target platforms for real-time video processing, paving the way for more complex algorithms and applications.

We have proposed baseline parallel schemes and data layouts for the disparity estimation algorithms that follow general optimization rules based on a simple GPU performance model. They are designed to gracefully scale on the forthcoming GPU architectures, like NVIDIA Pascal. Then, we have optimized the baseline code and improved performance around 3 times with different specific strategies, like vectorization or $CTA$-$to$-$Warp$ conversion, that are also expected to be valid for forthcoming architectures.


We plan to prove the higher performance potential of the new embedded NVIDIA Pascal GPUs to enable real-time implementations with larger images and a higher number of disparity levels, and more complex algorithms that provide better estimation results. In this sense, we are going to include post-filtering steps such as Left-Right Consistency Check, subpixel calculation, and adaptive P2, which are well-known methods of increasing accuracy.


\vspace{-0.15cm}
\section*{Acknowledgements}
\vspace{-0.15cm}

\label{sect:acks}
This research has been supported by the MICINN under contract number TIN2014-53234-C2-1-R. By the MEC under contract number TRA2014-57088-C2-1-R, the spanish DGT project SPIP2014-01352, and the Generalitat de Catalunya projects 2014-SGR-1506 and 2014-SGR–1562.
We thank Nvidia for the donation of the systems used in this work.
\vspace{-0.15cm}

%
\label{sect:bib}
\bibliographystyle{abbrv}
\bibliography{easychair}

\end{document}